\documentclass{article}

\newcommand{\version}{\texttt{v1.0} -- 13 April 2020}
\usepackage{arxiv}

\usepackage[utf8]{inputenc} % allow utf-8 input
\usepackage[T1]{fontenc}    % use 8-bit T1 fonts
\usepackage{hyperref}       % hyperlinks
\usepackage{url}            % simple URL typesetting
\usepackage{booktabs}       % professional-quality tables
\usepackage{amsfonts}       % blackboard math symbols
\usepackage{nicefrac}       % compact symbols for 1/2, etc.
\usepackage{microtype}      % microtypography
\usepackage{lipsum}
\usepackage{amsmath}
\usepackage{amsthm}
\usepackage{tcolorbox}
\usepackage{comment}
\usepackage{centernot}

\newtheorem{definition}{Definition}[section]

\newcommand{\nemo}{\textit{NEMO}}

\title{Technical Report:\\ \nemo{} Quantization for Deployment Model}

\author{
  Francesco Conti \\
  \texttt{fconti@iis.ee.ethz.ch -- f.conti@unibo.it} \\
}

\begin{document}
\maketitle

\begin{abstract}
This technical report aims at defining a formal framework for Deep Neural Network layer-wise quantization, focusing in particular on the problems related to the final deployment.
It also acts as a documentation for the \nemo{} (NEural Minimization for pytOrch) framework.
It describes the four DNN representations used in NEMO (\textit{FullPrecision}, \textit{FakeQuantized}, \textit{QuantizedDeployable} and \textit{IntegerDeployable}), focusing in particular on a formal definition of the latter two.
 An important feature of this model, and in particular the \textit{IntegerDeployable} representation, is that it enables DNN inference using purely integers -- without resorting to real-valued numbers in any part of the computation and without relying on an explicit fixed-point numerical representation.
\end{abstract}

\section{\textit{FullPrecision} representation}
The \textit{FullPrecision} representation is simply the ``normal''  one for real-valued neural networks.
We build a layer of a Deep Neural Network (DNN) out of a composition of operators in the Linear, Batch-Normalization, Activation classes.
\textbf{Linear} operators include convolutions, fully-connected layers (i.e., tensorwise matrix multiplication).
\textbf{Batch-Normalization} operators are also linear or affine transformations, but we treat them separately.
Non-linear \textbf{Activation} layers include the ReLU activation and variants in normal DNNs.

We formally define a \textit{layer} as any linear sequence of operators that takes as input the output of another layer and concludes with the first Activation layer in the sequence.
Note that in our model, we disallow branches starting from a layer that is not a Activation layer.

\begin{tcolorbox}[colback=green!5!white,colframe=green!75!black,title=In \nemo{}...]
For all intents and purposes, a \textit{FullPrecision} representation in \nemo{} is simply a valid PyTorch DNN model respecting these restrictions.
\end{tcolorbox}

\subsection{Linear operators}
A Linear operator has the form
\begin{equation}
    \xi =\mathbf{b} + \varphi = \mathbf{b} + \langle \mathbf{w}, \mathbf{x} \rangle = \mathbf{b}  + \sum_n \mathbf{w}_n \cdot \mathbf{x}_n \; ,
    \label{eq:affine}
\end{equation}
where $\mathbf{w}$, $\mathbf{x}$ are two tensors of \textit{weights} and \textit{input activations}, $\langle\cdot,\cdot\rangle$ indicates an elementwise product followed by a reduction along some of the tensor dimensions (essentially a scalar product).
Often, it is possible to neglect the bias term $\mathbf{b}$ as this can be incorporated in one of the following operators. In that case,
\begin{equation}
    \varphi = \langle \mathbf{w}, \mathbf{x} \rangle = \sum_n \mathbf{w}_n \cdot \mathbf{x}_n \; .
    \label{eq:linear}
\end{equation}
$\mathbf{w}$ is always at least 2-dimensional, indicating a mapping between $\mathcal{N}_{ic}$ input channels and $\mathcal{N}_{oc}$ output channels; $\mathbf{x}$ is always at least 1-dimensional, with $\mathcal{N}_{ic}$ input channels\footnote{
This is the case of fully-connected operators, where ``channels'' are constituted by a single element, and are sometimes called neurons.
}.
As a consequence, $\varphi$ has at least 1 dimension, $\mathcal{N}_{oc}$ output channels.

\subsection{Batch-Normalization operators}
Linear operators may be followed by Batch-Normalization (BN) operators.
BN acts a further affine transformation applied on $\varphi$ using parameters extracted statistically during training ($\mu$,$\sigma$) or trained with backpropagation ($\gamma$, $\beta$):
\begin{equation}
    \phi = \frac{\gamma}{\sigma} (\varphi-\mu) + \beta \; .
    \label{eq:bn}
\end{equation}
All BN parameters have only one dimension $\mathcal{N}_{oc}$.

\subsubsection{Activation operators}
Non-linear activations operate pointwise, as such their output $\mathbf{y}$ is dimensionally identical to $\phi$. They have the form:
\begin{equation}
    \mathbf{y} = {\textsc{Act}} (\phi) \; .
    \label{eq:act}
\end{equation}
The most common activation is $\textsc{ReLU}$:
\begin{equation}
    \mathbf{y} = {\textsc{ReLU}} (\phi) = \mathrm{clip}_{[0,+\infty)}(\phi) \; .
    \label{eq:relu_act}
\end{equation}

\section{\textit{FakeQuantized} representation}

In this Section, we discusse the \textit{FakeQuantized} representation of \nemo{}, that is used to represent a DNN in a form that takes quantization into account, but is still entirely manageable both in terms of topological transformations and training.
We start by formally defining what we mean by ``quantization'' in this document, along with a set of related definitions.

\subsection{Formal definition of DNN tensor quantization}

\begin{tcolorbox}[colback=blue!5!white,colframe=blue!75!black]
\begin{definition}
We call \textbf{quantized} a tensor $\mathbf{t}$ where all elements $t_i\in\mathbf{t}$ can be written as
\begin{equation}
    t_i = \alpha_\mathbf{t} + \varepsilon_\mathbf{t}\cdot {q_i}, \; {q_i}\in\mathbb{Z}_\mathbf{t} \; ,
    \label{eq:4}
\end{equation}
where $\varepsilon_\mathbf{t}$ is a scalar number in $\mathbb{R}$, which we call \textbf{quantum}\footnote{In this document we refer to layer-wise quantization. For channel-wise quantization, $\mathbf{\varepsilon}_\mathbf{t}$ is a vector of size $\mathcal{N}_{c}$.}
, $\alpha_\mathbf{t}$ is a scalar in $\mathbb{R}$ called \textbf{offset}, and $\mathbb{Z}_\mathbf{t}$ is a finite subset of $\mathbb{Z}$, which we call \textbf{quantized space}.
\end{definition}
\end{tcolorbox}

Therefore, the problem of \textit{quantization} of a DNN is that of defining a mapping of all the fundamental tensors of a DNN layer ($\mathbf{w}$, $\mathbf{x}$, $\mathbf{b}$, $\mathbf{y}$) to quantized tensors.
Considering that the ``natural'' representation of these tensors is real-valued ($\mathbf{t}\in\mathbb{R}$) (and practically implemented using 32-bit floating point numbers), a reasonable approach is to define a function $\textsc{q}$ to map $\mathbb{R}\rightarrow\mathbb{Z}_\mathbf{t}$ and combine it with Eq.~\ref{eq:4}.
This leads to the following definition:
\begin{tcolorbox}[colback=blue!5!white,colframe=blue!75!black]
\begin{definition}
We call \textbf{quantized version} of $\mathbf{t}$ a tensor $\widehat{\mathbf{t}}$ such that
\begin{equation}
    \widehat{\mathbf{t}} = \alpha_\mathbf{t} + \varepsilon_\mathbf{t}\cdot\textsc{q}_\mathbf{t}(\mathbf{t}) \; ,
\end{equation}
where $\textsc{q}_\mathbf{t}: \mathbb{R}^\mathcal{D}\rightarrow\mathbb{Z}_\mathbf{t}^\mathcal{D}$ ($\mathcal{D}$ being the dimensionality of $\mathbf{t}$) is a mapping from real to integer numbers that is pointwise, monotonic and piecewise constant, called the \textbf{quantization function}.
We call $\textsc{q}_\mathbf{t}(\mathbf{t})$ the \textbf{integer image} of $\mathbf{t}$.
\label{def:integer_image}
\end{definition}
\end{tcolorbox}

\subsection{Quantization-aware training}

The final objective of quantizing a DNN is using $\textsc{q}_\mathbf{t}({\mathbf{t})}$ in place of ${\mathbf{t}}$ without dropping accuracy.
This is targeted primarily by tuning the $\textsc{q}_\mathbf{t}$ functions used for the various tensors in a layer, and is currently the objective of extensive research.
The smaller is the cardinality of $\mathbb{Z}_\mathbf{t}$, $\mathcal{C}(\mathbb{Z}_\mathbf{t})$, the smaller will be the number of bits necessary to represent it in a hardware or software implementation.

\begin{tcolorbox}[colback=green!5!white,colframe=green!75!black,title=In \nemo{}...]
A \textit{FakeQuantized} representation is one that imposes that the weights of Linear operators and the output of Activation operators are real valued, but chosen from a restricted set of \textit{quantized} values during forward-propagation.
Note that this restriction is not usually applied to other layers.
This version of the network \texttt{net} can be obtained by running
\begin{verbatim}
  net = nemo.transform.quantize_pact(net, dummy_input=dummy_input)
\end{verbatim}
where \texttt{dummy\_input} is a \texttt{torch.Tensor} sized like the network input.
Currently, nemo supports a PACT-like~\cite{ChoiPACTParameterizedClipping2018} linear quantization scheme for both weights and activations.
\end{tcolorbox}

In the example case of a ReLU Activation using PACT~\cite{ChoiPACTParameterizedClipping2018}, this means that the activation is replaced with
$$
    \mathbf{y} = \mathrm{ReLU}(\mathbf{\phi}) = \mathrm{clip}_{ [0,\infty) }(\mathbf{\phi}) \quad\longrightarrow\quad \mathbf{y} = \left\lfloor 1/\varepsilon_\mathbf{y} \cdot \mathrm{clip}_{ [0,\beta_\mathbf{y}) } (\mathbf{\phi}) \right\rfloor \cdot \varepsilon_\mathbf{y} \; .
$$
Two changes are introduced to the ReLU.
First, the clipping function is not only clipping at 0, but also at a maximum value $\beta$, which can be set to the maximum value of $\mathbf{y}$ in the \textit{FullPrecision} stage (see later).
Second, the Activation explicitly uses the \textit{quantum} inside.
To represent the tensor $\mathbf{y}$ with $Q$ bits, $\varepsilon_\mathbf{y} = \beta_\mathbf{y} / (2^Q - 1)$.
Due to the clipping nature of ReLUs, we set $\alpha_\mathbf{y} = 0$ for all activations.

\begin{tcolorbox}[colback=green!5!white,colframe=green!75!black,title=In \nemo{}...]
For historical reasons, in \texttt{PACT\_Act} activations the parameter we call $\beta$ in this document is saved in the \texttt{alpha} parameter.
This may change in future versions!
\end{tcolorbox}

Linear weights are stored in full-precision, but a similar clipping function is used at runtime in forward-propagation (when using linear PACT-like quantization):
$$
\widehat{\mathbf{w}} = \left\lfloor 1/\varepsilon_\mathbf{w} \cdot \mathrm{clip}_{ [\alpha_\mathbf{w},\beta_\mathbf{w}) } (\mathbf{\mathbf{w}}) \right\rfloor \cdot \varepsilon_\mathbf{w} \; .
$$
$\widehat{\mathbf{w}}$ is used in place of $\mathbf{w}$ when performing forward-propagation.

\begin{tcolorbox}[colback=green!5!white,colframe=green!75!black,title=In \nemo{}...]
For historical reasons, in \texttt{PACT\_Conv2d} and other Linear layers activations the parameter we call $\alpha$ in this document is saved in the \texttt{alpha} parameter with inverted sign (so it's typically \textit{positive}, because weights are usually zero-crossing).
This may change in future versions!
\end{tcolorbox}

To enable training of the network, quantization-aware training strategies replace tensors with their quantized version only during the forward-propagation step, but they use and update real tensors in backward-propagation.
Most methods estimate gradients through non-linear quantization functions using the \textit{straight-through estimator} (STE), i.e., they simply work on full-precision tensors ignoring all quantization functions~\cite{ChoiPACTParameterizedClipping2018}.
The fundamentals behind the fact that STE works are only recently being understood (see Spallanzani~et~al.~\cite{SpallanzaniAdditiveNoiseAnnealing2019}).

\begin{tcolorbox}[colback=green!5!white,colframe=green!75!black,title=In \nemo{}...]
We use PACT-like quantization for both activations and weights, which employs the STE.
Therefore, if $\mathcal{L}$ is the loss, for activations:
\begin{align*}
\mathbf{\nabla}_\mathbf{\phi} \mathcal{L} &\doteq \chi_{[0,\beta_\mathbf{y})}(\phi) \cdot \mathbf{\nabla}_\mathbf{y} \mathcal{L} 
\end{align*}
and for weights:
$$
\mathbf{\nabla}_\mathbf{{w}} \mathcal{L} \doteq \chi_{[\alpha_\mathbf{w},\beta_\mathbf{w})}(\mathbf{w}) \cdot \mathbf{\nabla}_\mathbf{\varphi} \mathcal{L}
$$
Both the forward- and backward-prop functions are defined in the same \texttt{nemo.quant.pact.PACT\_QuantFunc} and \texttt{nemo.quant.pact.PACT\_QuantFunc\_Asymm} \texttt{torch.autograd.Function}s for activations and weights, respectively.
\end{tcolorbox}

\section{\textit{QuantizedDeployable} and \textit{IntegerDeployable} representations}

While the \textit{FakeQuantized} representation is useful for training and quantization-aware fine-tuning, it  cannot directly be used  for deployment on an integer-only Quantized Neural Network (QNN), because quantization is defined rigorously only for weights and activations, but not for all the intermediate representations.

The \textit{QuantizedDeployable} representations ``completes'' the task started by the \textit{FakeQuantized} transformation: all operators on the network operate on quantized inputs and produce quantized outputs.
Since all quantized tensors have an integer image as defined in Definition~\ref{def:integer_image}, it is possible to completely get rid of their real-valued nature and use only integer images along the network.
This step yields a \textit{IntegerDeployable} representation.
In this Section, we describe simultaneously the \textit{QuantizedDeployable} and \textit{IntegerDeployable} representations, as they are one the image of the other through Definition~\ref{def:integer_image}.

\begin{tcolorbox}[colback=green!5!white,colframe=green!75!black,title=In \nemo{}...]
Transforming a model \texttt{net} into \textit{QuantizedDeployable} representation requires three distinct operations.
First, quantizing BatchNormalization layers (see Section~\ref{sec:quant_bn}): 
\begin{verbatim}
  net = nemo.transform.bn_quantizer(net)
\end{verbatim}
Second, freezing Linear weights in their quantized state (i.e., setting $\mathbf{w}\leftarrow\widehat{\mathbf{w}}$):
\begin{verbatim}
  net.harden_weights()
\end{verbatim}
Third, propagating $\varepsilon$ quanta along the network, as explained in detail for each operator in all parts of this Section:
\begin{verbatim}
  net.set_deployment(eps_in=1./255)
\end{verbatim}
To switch to \textit{IntegerDeployable}, several operators have to be changed and all parameters have to be replaced by integer ones:
\begin{verbatim}
  net = nemo.transform.integerize_pact(net, eps_in=1.0/255)
\end{verbatim}
Note that in all representations, \nemo{} utilizes \texttt{float32} to represent data.
This means that \nemo{} networks in \texttt{IntegerDeployable} format can be inferred on a GPU with no efficient integer support paying only a small penalty because of the additional operators discussed in this section.
\end{tcolorbox}

\subsection{Quantization/Activation operators}
From the simple consideration that the input of a DNN layer typically comes from the output of another layer, follows that a favourable location to place the quantization function for activation tensors is within the activation operator, which produces the input to the next block.
There is another fundamental consideration that singles out this operator as the right one for embedding the quantization function: $\textsc{q}$ is by construction non-linear and clipped, both characteristics shared with ReLU (which is clipped only on the lower side) and other activation operators (most of which are clipped on both sides).

The \textit{Quantization/Activation} operator, in this case, provides the double functionality of \textit{i)} being the non-linear activation essential for the DNN to work; \textit{ii)} squashing the input tensor $\mathbf{t}$ (which might be real or quantized within its own quantized space $\mathbb{Z}_\mathbf{t}$) into a (generally smaller) quantization space $\mathbb{Z}_\mathbf{y}$.
Therefore, whereas the quantization function as defined in Eq.~\ref{eq:4} is parametrized to the same quantization space to which it is applied (e.g., $\textsc{q}_\mathbf{t}(\mathbf{t})$), the quantization/activation function is parametrized to the target quantization space $\mathbb{Z}_\mathbf{y}$ (e.g., $\textsc{q}_\mathbf{y}(\mathbf{t})$).

\paragraph{General case of quantization functions.}
To understand in depth how a quantization function works, we start from the explicit mapping of a real-valued tensor $\mathbf{t}$ to an \textit{arbitrarily defined} integer image.
By Definition~\ref{def:integer_image}, this function is a ladder mapping the input tensor to the integer image of the target tensor:
\begin{equation}
	\textsc{q}_\mathbf{y}(\mathbf{t}) = \sum_{i=M}^{N-1} i\cdot\chi_{[\tau_{i},\eta_{i})} (\mathbf{t}) , \;\;\;\; \tau_i < \eta_i \leq \tau_{i+1}, \;\forall i \label{eq:quant}
\end{equation}
where  $\tau_i$ and $\eta_i$ are a set of threholds identifying the interval of $\mathbb{R}$ mapped to each value $z\in\mathbb{Z}_\mathbf{y}$; $M,N$ are the lower and upper value of $\mathbb{Z}_\mathbf{y}$, respectively.
Here we focus on quantization functions that are continuously defined: they set $\eta_i=\tau_{i+1}$ to represent a continuous interval of $\mathbb{R}$ and they set them along a continuous function $\tau_i = \tau(i)$ mapping $\mathbb{N}\rightarrow\mathbb{R}$.

The quantization function does not need to be applied to a real-valued tensor, but can be applied directly on its integer image:
\begin{align*}
	\textsc{q}_\mathbf{y}(\mathbf{t}) = \sum_{i=M}^{N-1} i\cdot\chi_{[\widehat\tau_{i},\widehat\eta_{i})} \Big(\textsc{q}_\mathbf{t}(\mathbf{t})\Big) \;\;,\;\;
	\widehat\tau_i = \left\lceil \frac{\tau_i}{\varepsilon_\mathbf{t}} \right\rceil \;\;,\;\;
	\widehat\eta_i = \left\lceil \frac{\eta_i}{\varepsilon_\mathbf{t}} \right\rceil 
\end{align*}
By changing indeces, and defining an $\alpha_\mathbf{y}$, $P=N-M$, it is possible to have the staircase always starting from index 0, which gives a ``canonical'' form of quantization function:
\begin{equation}
	\textsc{q}_\mathbf{y}(\mathbf{t}) = \alpha_\mathbf{y}+ \sum_{j=0}^{P-1} j\cdot\chi_{[\widehat\tau_{j},\widehat\eta_{j})} \Big(\textsc{q}_\mathbf{t}(\mathbf{t})\Big)
	\label{eq:iquant}
\end{equation}

\paragraph{Linear quantization.}
Linear quantization uses an affine transformation to derive $\tau_i$ from $i$; this translates the abstract formulation of Eq.~\ref{eq:quant} to a clip function, which is what was shown without full explanation in previous Sections:
\begin{equation}
	\textsc{lq}_\mathbf{y}(\mathbf{t}) =
	 \frac{\alpha_\mathbf{y}}{\varepsilon_\mathbf{y}} + \sum_{i=0}^{N-1}i\cdot \chi_{[i,i+1)} \left(\frac{\mathbf{t}-\alpha_\mathbf{y}}{\varepsilon_\mathbf{y}}\right) = 
	\mathrm{clip}_{[\alpha_\mathbf{y}/\varepsilon_\mathbf{y},\beta/\varepsilon_\mathbf{y})}
	\left( \left\lfloor \frac{\mathbf{t}}{\varepsilon_\mathbf{y}} \right\rfloor \right)
	\label{eq:lq_quant_real}
\end{equation}
with $N=\nicefrac{\beta_\mathbf{y}-\alpha_\mathbf{y}}{\varepsilon_\mathbf{y}}$. 

How to perform this operation when starting from an integer image?
One possibility is to directly apply Eq.~\ref{eq:iquant}, which translates on a comparison with a set of explicitly defined thresholds.
This approach might be expensive to perform in an actual deployment, but it requires no approximation.
See also Section~\ref{sec:quant_bn} for a practical case where we follow this route.

The alternative relies on a technique that we call \textit{requantization}: this requires an approximation and is the object of the following section.
Here we anticipate the final result in this case:
\begin{equation}
	\textsc{lq}_\mathbf{y}(\mathbf{t}) \approx 
    \mathrm{clip}_{[\alpha/\varepsilon_\mathbf{y},\beta/\varepsilon_\mathbf{y})}\left( \left\lfloor\frac{\varepsilon_\mathbf{t}\cdot 2^d}{\varepsilon_\mathbf{y}}\right\rfloor \cdot \textsc{q}_\mathbf{t}(\mathbf{t}) \gg d \right) \label{eq:lq_scaled}
\end{equation}
where $d$ is an appropriately chosen integer (see Section~\ref{sec:requantization}).

\begin{tcolorbox}[colback=green!5!white,colframe=green!75!black,title=In \nemo{}...]
When switching to \textit{QuantizedDeployable} representation, \texttt{nemo.quant.pact.PACT\_Act} activations use the ``regular'' definition of Eq.~\ref{eq:lq_quant_real}.
\end{tcolorbox}
\begin{tcolorbox}[colback=green!5!white,colframe=green!75!black,title=In \nemo{}...]
In \textit{IntegerDeployable} representation, \texttt{nemo.quant.pact.PACT\_Act}s are transformed into \texttt{nemo.quant.pact.PACT\_IntegerAct} activations, which apply the requantization method presented in~Eq.~\ref{eq:lq_scaled}.
\end{tcolorbox}

\subsection{Requantization}
\label{sec:requantization}

The \textit{requantization} function is essential in any case where we have to transform a tensor from one quantized space to a different one.
Ideally, this would happen by simply scaling the quanta:
$$
	\textsc{q}_\mathbf{b}(\mathbf{a}) \leftarrow  \frac{\varepsilon_\mathbf{a}}{\varepsilon_\mathbf{b}} \cdot \textsc{q}_\mathbf{a}(\mathbf{a})
$$
In general $\varepsilon_\mathbf{a}/\varepsilon_\mathbf{b}$ is not an integer, and so this function cannot be used to define an integer image $\textsc{q}_\mathbf{b}(\mathbf{a})$. 
To solve this issue with an approximation, let us introduce an arbitrary natural number $D$. Then, we can express the ratio as a limit:
$$
\frac{\varepsilon_\mathbf{a}}{\varepsilon_\mathbf{b}} =
\lim_{D\rightarrow\infty} \left\lfloor\frac{\varepsilon_\mathbf{a}\cdot D}{\varepsilon_\mathbf{b}}\right\rfloor \cdot \frac{1}{D}
$$
While $D$ cannot be infinite in practice, this suggests we can make it arbitrarily big to reduce the error in the ratio as much as possible.
What is the error in that case? By definition of the floor function,
\begin{align*}
\frac{\varepsilon_\mathbf{a}\cdot D}{\varepsilon_\mathbf{b}} -
 \left\lfloor\frac{\varepsilon_\mathbf{a}\cdot D}{\varepsilon_\mathbf{b}}\right\rfloor &< 1 \;\implies\;
  \frac{\varepsilon_\mathbf{a}}{\varepsilon_\mathbf{b}} -
 \left\lfloor\frac{\varepsilon_\mathbf{a}\cdot D}{\varepsilon_\mathbf{b}}\right\rfloor \cdot \frac{1}{D} < \frac{1}{D}
\end{align*}
therefore the error is bound by $1/D$.
To limit the relative error to less than a fraction $\eta$, then,
$$
\frac{1/D}{\varepsilon_\mathbf{a}/\varepsilon_\mathbf{b}} = \frac{\varepsilon_\mathbf{b}}{\varepsilon_\mathbf{a}\cdot D} \leq \eta \; \implies \; D \geq \frac{\varepsilon_\mathbf{b}}{\varepsilon_\mathbf{a}\cdot\eta}
$$

Let us use this concept for a formal definition of the requantization function:
\begin{tcolorbox}[colback=blue!5!white,colframe=blue!75!black]
\begin{definition}
Let us consider two quantized spaces $\mathbb{Z}_\mathbf{a}$, $\mathbb{Z}_\mathbf{b}$, their related quanta $\varepsilon_\mathbf{a}$, $\varepsilon_\mathbf{b}$, and an integer image $\textsc{q}_\mathbf{a}(\mathbf{a})$ in the first quantized space. 
We define the \textbf{requantization} function from $\mathbb{Z}_\mathbf{a}$ to $\mathbb{Z}_\mathbf{b}$ as
\begin{equation}
	\textsc{rq}_{(\mathbb{Z}_\mathbf{a}\rightarrow\mathbb{Z}
	_\mathbf{b},D)}\Big(\textsc{q}_\mathbf{a}(\mathbf{a})\Big) = \left\lfloor\frac{\varepsilon_\mathbf{a}\cdot D}{\varepsilon_\mathbf{b}}\right\rfloor \cdot \frac{\textsc{q}_\mathbf{a}(\mathbf{a})}{D}
	\label{eq:requantization}
\end{equation}
where $D\in\mathbb{N}$ is a parameter chosen arbitrarily.
\end{definition}
\end{tcolorbox}

Under this definition, we can approximate the integer image of tensor $\mathbf{a}$ in the quantized space $\mathbb{Z}_\mathbf{b}$ as 
$$
\textsc{q}_\mathbf{b}(\mathbf{a}) \approx
	\textsc{rq}_{(\mathbb{Z}_\mathbf{a}\rightarrow\mathbb{Z}
	_\mathbf{b},D)}\Big(\textsc{q}_\mathbf{a}(\mathbf{a})\Big)\; .
$$
We typically choose $D=2^d$ as a power of 2. In this way, the division reduces to a right shift:
\begin{equation}
	\textsc{rq}_{(\mathbb{Z}_\mathbf{a}\rightarrow\mathbb{Z}
	_\mathbf{b},D)}\Big(\textsc{q}_\mathbf{a}(\mathbf{a})\Big) = \left\lfloor\frac{\varepsilon_\mathbf{a}\cdot 2^d}{\varepsilon_\mathbf{b}}\right\rfloor \cdot {\textsc{q}_\mathbf{a}(\mathbf{a})} \gg d \; ,
	\label{eq:requantization_pow2}
\end{equation}
and the $d$ parameter can be bound to a relative error $\eta$ with
\begin{equation}
	d \geq \log_2 \frac{\varepsilon_\mathbf{b}}{\varepsilon_\mathbf{a}\cdot\eta} \; .
	\label{eq:requantization_error}
\end{equation}

The requantization approximation can be used to derive the linear quantization transformation presented without proof in the previous section.
\begin{align*}
	\textsc{lq}_\mathbf{y}(\mathbf{t}) &= \frac{\alpha_\mathbf{y}}{\varepsilon_\mathbf{y}} +
	\mathrm{clip}_{[0,\beta_\mathbf{y}/\varepsilon_\mathbf{y}-\alpha_\mathbf{y}/\varepsilon_\mathbf{y})}
	\left( \left\lfloor \frac{\mathbf{t}-\alpha_\mathbf{y}}{\varepsilon_\mathbf{y}} \right\rfloor \right) =\\
	&= \frac{\alpha_\mathbf{y}}{\varepsilon_\mathbf{y}} + \mathrm{clip}_{[0,\beta_\mathbf{y}/\varepsilon_\mathbf{y}-\alpha_\mathbf{y}/\varepsilon_\mathbf{y})}
	\left( \left\lfloor \frac{\varepsilon_\mathbf{t}\cdot\textsc{q}_\mathbf{t}(\mathbf{t})}{\varepsilon_\mathbf{y}} \right\rfloor \right) \approx \\
	&\approx \frac{\alpha_\mathbf{y}}{\varepsilon_\mathbf{y}} + \mathrm{clip}_{[0,\beta_\mathbf{y}/\varepsilon_\mathbf{y}-\alpha_\mathbf{y}/\varepsilon_\mathbf{y})}
	\left( \left\lfloor \left\lfloor \frac{\varepsilon_\mathbf{t}}{\varepsilon_\mathbf{y}}\cdot 2^d \right\rfloor \frac{\textsc{q}_\mathbf{t}(\mathbf{t})}{2^d} \right\rfloor \right) = \\ 
	\textsc{lq}_\mathbf{y}(\mathbf{t}) 
	&= \frac{\alpha_\mathbf{y}}{\varepsilon_\mathbf{y}} + \mathrm{clip}_{[0,\beta_\mathbf{y}/\varepsilon_\mathbf{y}-\alpha_\mathbf{y}/\varepsilon_\mathbf{y})}
	\left( \left\lfloor \frac{\varepsilon_\mathbf{t}\cdot 2^d}{\varepsilon_\mathbf{y}} \right\rfloor\cdot {\textsc{q}_\mathbf{t}(\mathbf{t})}\gg d \right) \; ,
\end{align*}
which is an alternative form of Eq.~\ref{eq:lq_scaled}.

\begin{tcolorbox}[colback=green!5!white,colframe=green!75!black,title=In \nemo{}...]
\texttt{nemo.quant.pact.PACT\_IntegerAct} activations used in the \textit{IntegerDeployable} representation compute $d$ internally. They use Eq.~\ref{eq:requantization_error} given an attribute called \texttt{requantization\_factor}, that is $1/\eta$ and defaults to 16.
\end{tcolorbox}

\subsection{Linear operators}
Let us now assume that $\widehat{\mathbf{w}}\in[\alpha_\mathbf{w}, \beta_\mathbf{w})$, $\widehat{\mathbf{x}}\in[0,\beta_\mathbf{x})$ are the quantized versions of $\mathbf{w}$, $\mathbf{x}$. 
Following Eq.~\ref{eq:linear}, approximating a linear layer by using quantized versions of $\mathbf{w}$, $\mathbf{x}$ means the following
\begin{equation}
  \begin{aligned}
    \widehat{\varphi} &= \sum_n \widehat{\mathbf{w}_n} \cdot \widehat{\mathbf{x}_n} = \\
       &= \alpha_\mathbf{w}\varepsilon_\mathbf{x}\sum_n\textsc{q}_\mathbf{x}(\mathbf{x}_n) + \varepsilon_\mathbf{w}\varepsilon_\mathbf{x}\sum_n\textsc{q}_\mathbf{w}(\mathbf{w}_n)\cdot \textsc{q}_\mathbf{x}(\mathbf{x}_n)\\
       &\doteq \alpha_\varphi + \varepsilon_\varphi \cdot \textsc{q}(\varphi)
  \end{aligned}
\end{equation}
neglecting the bias term.
$\widehat\varphi$ is \textit{not} explicitly defined as the quantized version of $\varphi$; however, it is still a quantized tensor, where the quantum is $\varepsilon_\varphi = \varepsilon_\mathbf{w}\cdot\varepsilon_\mathbf{x}$, and the integer image is
\begin{equation}
    \textsc{q}(\varphi) = \sum_n\textsc{q}_\mathbf{w}(\mathbf{w}_n)\cdot \textsc{q}_\mathbf{x}(\mathbf{x}_n) \label{eq:ilinear}
\end{equation}
As a consequence, the quantization space of $\varphi$ is given by
\begin{equation}
    \mathbb{Z}_\varphi = \left\{ z_\varphi : z_\varphi = n\cdot z_\mathbf{w} \cdot z_\mathbf{x}\right\} , \, 0\leq n<N, z_\mathbf{w}\in\mathbb{Z}_\mathbf{w},\, z_\mathbf{x}\in\mathbb{Z}_\mathbf{x}
\end{equation}
In a practical implementation $\widehat\varphi$ will have to be represented with a larger number of bits than $\mathbf{w}$, $\mathbf{x}$.

Note that nothing directly guarantees that $\widehat\varphi$ is a \textit{good} approximation of $\varphi$.
However, if the network has been trained/fine-tuned in \textit{FakeQuantized} representation, it is not really important to approximate $\varphi$: $\widehat\varphi$ was actually used in forward-prop training, not $\varphi$!
In practice, for not too strong quantizations, \textit{FakeQuantized} fine-tuning might not even be necessary. A simple validation will verify that $\widehat\varphi$ propagates the correct information through the network.

\begin{tcolorbox}[colback=green!5!white,colframe=green!75!black,title=In \nemo{}...]
The behavior of Linear operators such as \texttt{PACT\_Conv2d} does not change from \textit{FakeQuantized} to \textit{QuantizedDeployable}.
The \texttt{net.harden\_weights()} call replaces all weights ${\mathbf{w}}$ with their quantized version $\widehat{\mathbf{w}}$.
The quantum $\varepsilon_\varphi$ after the Linear operation is computed automatically  by \nemo{}.
\end{tcolorbox}
\begin{tcolorbox}[colback=green!5!white,colframe=green!75!black,title=In \nemo{}...]
In \textit{IntegerDeployable} representation, the operator also works in the same way, but the \texttt{nemo.transform.integerize\_pact} function will replace all weights $\widehat{\mathbf{w}}$ with their integer image $\textsc{q}_\mathbf{w}(\mathbf{w})$.
\end{tcolorbox}

\subsection{Batch-Normalization operators}
Equation~\ref{eq:bn} involves an affine transformation with parameters ($\gamma$,$\beta$,$\mu$,$\sigma$) that are, in general, in the real domain.
Batch-Normalization is often very important for quantization strategy,  it normalizes activations, constraining  ``softly'' in an interval that maps well to the clipping ($\beta$) that is imposed through quantization.
In general, three different strategies can be applied: \textit{i)} \textit{fold} the network BN operators in the previous linear operator, before performing its quantization; \textit{ii)} replace the parameters with quantized versions; \textit{iii)} merge the BN operator with the following activation function, creating appropriate thresholds.

\paragraph{BN Folding.}
Integrating Eq.~\ref{eq:bn} with Eq.~\ref{eq:affine},
\begin{align*}
	\phi &= \gamma/\sigma\sum_n\mathbf{w}_n\cdot\mathbf{x}_n-\mu\gamma/\sigma+\beta = \\
		&= \sum_n \frac{\gamma}{\sigma}\mathbf{w}_n\cdot\mathbf{x}_n - \mu\gamma/\sigma + \beta \doteq\\
		&\doteq \sum_n \mathbf{w'}_n\cdot \mathbf{x}_n + \mathbf{b'}
\end{align*}
Therefore, folding a BN layer into the linear layer that precedes it involves replacing its parameters with the following transform:
\begin{equation}\begin{aligned}
	\mathbf{w} &\longleftarrow \gamma / \sigma \cdot \mathbf{w} \\
	\mathbf{b} &\longleftarrow \mathbf{b} + \beta - \gamma / \sigma \cdot \mu
\end{aligned}\end{equation}
Note that even if the original linear layer had no bias term, the folded linear layer in general will have a bias to take into account the affine transformation in the BN layer.

\begin{tcolorbox}[colback=green!5!white,colframe=green!75!black,title=In \nemo{}...]
BN folding of a model net can be performed at the \textit{FakeQuantization} stage by calling
\begin{verbatim}
  net.fold_bn()
  net.reset_alpha_weights()
\end{verbatim}
with an optional dictionary of specific operators to be folded (the default is to fold all).
The second command is necessary to reset the $\alpha, \beta$ parameters of the weights after folding.
\end{tcolorbox}

\paragraph{Merging BN with Quantization/Activation.}
\label{sec:quant_bn}
An alternative way to remove a BN layer with respect to folding it into a convolution is to merge it with the following quantization/activation function, i.e., folding the affine transformation into the $\tau$ thresholds shown in Eq.~\ref{eq:quant}.

In the case of linear quantization (of all kinds), the procedure is particularly interesting and useful, as it can be used to absorb all real parameters without any approximation into a set of integer thresholds:
\begin{equation}
	\textsc{th}_i =\left\lceil \frac{1}{\varepsilon_\varphi}\left(\sigma/\gamma\cdot i \cdot \varepsilon_\mathbf{y} - \beta\sigma/\gamma + \mu \right) \right\rceil \label{eq:quant_th_thresholds}
\end{equation}
These thresholds map directly the integer image of $\phi$ to that of the output $\mathbf{y}$, therefore enabling execution of the layer entirely in the integer domain:
\begin{equation}
	\textsc{q}_\mathbf{y}(\varphi) = \sum_{i=0}^{N-1} i\cdot\chi_{[\textsc{th}_{i},\textsc{th}_{i+1})} \Big(\textsc{q}_\varphi(\varphi)\Big)  \label{eq:quant_th_act}
\end{equation}

{\small
\paragraph{Proof.} Propagating Eq.~\ref{eq:bn} means 
\begin{align*}
	{\textsc{lq}}(\widehat\phi) &= {\textsc{lq}}\Big( {\gamma/\sigma(\widehat\varphi-\mu)+\beta} \Big) = \\
	 &= \sum_{i=0}^{N-1} i\cdot \chi_{[i,i+1)} \left(\frac{\gamma/\sigma (\widehat\varphi -\mu)+\beta}{\varepsilon_\mathbf{y}}\right)
\end{align*}
Each element in the sum identified by index $i$ is non-zero if and only if
\begin{align*}
&i \leq \frac{\gamma/\sigma(\widehat\varphi-\mu)+\beta}{\varepsilon_\mathbf{y}}  < i+1 \\
&i\varepsilon_\mathbf{y} \leq \gamma/\sigma(\widehat\varphi-\mu)+\beta  < (i+1)\varepsilon_\mathbf{y}
\end{align*}
By construction or simple transformations, we can safely assume that $\gamma,\sigma>0$. Therefore the condition can be transformed in
\begin{equation*}\begin{aligned}
	\begin{cases}
		\widehat\varphi \geq \sigma/\gamma \cdot i     \cdot\varepsilon_\mathbf{y} - \beta\sigma/\gamma + \mu &\doteq \tau_i\\
		\widehat\varphi <    \sigma/\gamma \cdot (i+1) \cdot\varepsilon_\mathbf{y} - \beta\sigma/\gamma + \mu &\doteq \tau_{i+1}
	\end{cases}
\end{aligned}\end{equation*}
By Eq.~\ref{eq:ilinear}, $\widehat\varphi=\varepsilon_\varphi\cdot\textsc{q}(\varphi)$, therefore this is equivalent to
\begin{equation*}\begin{aligned}
	\begin{cases}
		\textsc{q}_\mathbf{y} (\varphi) \geq 1/{\varepsilon_\varphi}\cdot \left (\sigma/\gamma \cdot i     \cdot\varepsilon_\mathbf{y} - \beta\sigma/\gamma + \mu\right) \\ 
		\textsc{q}_\mathbf{y} (\varphi) < 1/{\varepsilon_\varphi}\cdot \left (\sigma/\gamma \cdot (i+1) \cdot\varepsilon_\mathbf{y} - \beta\sigma/\gamma + \mu\right) 
	\end{cases}
\end{aligned}\end{equation*}
Finally, as $\textsc{q}_\mathbf{y}(\varphi)$ is integer, one can define a set of integer thresholds absorbing all real parameters without any further approximation:
\begin{equation*}
	\textsc{th}_i =\left\lceil \frac{1}{\varepsilon_\varphi}\left(\sigma/\gamma\cdot i \cdot \varepsilon_\mathbf{y} - \beta\sigma/\gamma + \mu \right) \right\rceil 
\end{equation*}
corresponding to the complete quantization function:
\begin{equation*}
	\textsc{q}_\mathbf{y}(\varphi) = \sum_{i=0}^{N-1} i\cdot\chi_{[\textsc{th}_{i},\textsc{th}_{i+1})} \Big(\textsc{q}_\varphi(\varphi)\Big)  
\end{equation*}
\qed
}

The threshold-based approach is naturally especially effective when the number of thresholds is small, i.e. when the cardinality of $\mathbb{Z}_\mathbf{y}$ is small.

\begin{tcolorbox}[colback=green!5!white,colframe=green!75!black,title=In \nemo{}...]
While \nemo{} includes a threshold-based \texttt{nemo.quant.pact.PACT\_ThresholdAct} activation layer, its operation is experimental and unsupported in the current version.
\end{tcolorbox}

\paragraph{Integer BN.}
When the target cardinality of the output of a block ($\mathbb{Z}_\mathbf{y}$) is not particularly small, thresholds are not an efficient way to implement the BN and the quantization/activation; it is more effective to explicitly perform BN and then quantization/activation by means of Eq.~\ref{eq:lq_scaled}.
Executing the BN layer in the integer domain requires replacing the parameters of the BN with quantized versions (see Rusci~et~al.~\cite{RusciWorkinProgressQuantizedNNs2018,RusciMemoryDrivenMixedLow2019}), which means deriving a $\widehat\phi$ approximating $\phi$.
Here we consider $\widehat\varphi$ the ``correct'' input of which $\phi$ is a function.
Let $\kappa = \gamma/\sigma$, $\lambda = \beta - \kappa\cdot\mu$; then
\begin{align}
	\phi &= \gamma/\sigma \cdot\widehat\varphi-\gamma/\sigma\cdot\mu + \beta =  \notag \\
		&= \kappa \cdot \widehat\varphi + \lambda \approx\notag\\
	&\approx \widehat\kappa\cdot\widehat\varphi + \widehat\lambda \doteq \widehat\phi \label{eq:ibn1}
\end{align}
where $\widehat\kappa$ and $\widehat\lambda$ are the quantized versions of the respective parameters.
In general, $\lambda$ is represented in its own precision $\varepsilon_\lambda$ chosen independently, and then requantized  to $\varepsilon_\kappa\varepsilon_\varphi$ before using it.
Then, Eq.~\ref{eq:ibn1} becomes
\begin{align}
\widehat\phi &= \varepsilon_\kappa\varepsilon_\varphi \cdot \Big(\textsc{q}_\kappa(\kappa)\textsc{q}_\varphi(\varphi) + \textsc{rq}_{(\mathbb{Z}_\lambda\rightarrow\mathbb{Z}_\phi,D=1)}\big(\textsc{q}_\lambda(\lambda)\big)\Big) \doteq\notag \\
&= \varepsilon_\kappa\varepsilon_\varphi \cdot \big(\textsc{q}_\kappa(\kappa)\textsc{q}_\varphi(\varphi) + \textsc{q}_{\phi}(\lambda)\big) \doteq\notag \\
&\doteq \varepsilon_\phi \cdot \textsc{q}_\phi(\phi) \notag
\end{align}
Thus, in the domain of integer images,
\begin{equation}
	\textsc{q}_\phi(\phi) = \textsc{q}_\kappa(\kappa)\cdot\textsc{q}_\varphi(\varphi) + \textsc{q}_{\phi}(\lambda) \label{eq:ibn}
\end{equation}
Similarly to Eq.~\ref{eq:ilinear}, this allows to fully operate the BN layer in the integer domain of the integer images; the quantized space is
\begin{equation}
    \mathbb{Z}_\phi = \left\{ z_\phi : z_\phi = z_\kappa \cdot z_\varphi\right\} , \, z_\kappa\in\mathbb{Z}_\kappa,\, z_\varphi\in\mathbb{Z}_\varphi
\end{equation}

\begin{tcolorbox}[colback=green!5!white,colframe=green!75!black,title=In \nemo{}...]
In \textit{QuantizedDeployable} representation, \texttt{torch.nn.BatchNorm2d} is replaced with \texttt{nemo.quant.pact.PACT\_QuantizedBatchNorm2d}.
To quantize $\kappa$ and $\lambda$, we use a symmetric ($\alpha=-\beta$) $Q$-bit quantizer: we compute statically $\beta$ and set $\varepsilon=2\beta/(2^Q-1)$.
Requantization is not accurately represented at this representation level.
\end{tcolorbox}

\begin{tcolorbox}[colback=green!5!white,colframe=green!75!black,title=In \nemo{}...]
In \textit{IntegerDeployable} representation, \texttt{nemo.quant.pact.PACT\_QuantizedBatchNorm2d} is replaced with \texttt{nemo.quant.pact.PACT\_IntegerBatchNorm2d}.
$\textsc{q}_\lambda(\lambda)$ is requantized to $\mathbb{Z}_{\phi}$ before being used:
$$
\textsc{q}_{\phi}(\lambda) \doteq \textsc{rq}_{(\mathbb{Z}_\lambda\rightarrow\mathbb{Z}_\phi,D=1)}\big(\textsc{q}_\lambda(\lambda)\big)
$$
In this way, the choice whether to store $\lambda$ in a lower-precision format  $\textsc{q}_\lambda(\lambda)$ or directly in the target format  $\textsc{q}_\phi(\lambda)$ (which typically requires 32 bits) is left to the deployment backend.
$D=1$ ($d=0$) is currently wired.
\end{tcolorbox}

\subsection{Add operators}
When several paths in a DNN di-graph converge to the same node, they are typically combined through an Add operator.
An obvious requirement for these situations is that the numerical representations of each branch should be equalized to that of the others to be summable -- each tensor coming from a branch lives its own space $\mathbb{Z}_\mathbf{b0}$, $\mathbb{Z}_\mathbf{b1}$, $\dots$: therefore,
$$
\varepsilon_\mathbf{b0}\textsc{q}_{\mathbf{b0}}(\mathbf{b_0}) + \varepsilon_\mathbf{b1}\textsc{q}_{\mathbf{b1}}(\mathbf{b_1}) + \dots = \varepsilon_\mathbf{s}\textsc{q}_{\mathbf{s}}(\mathbf{s}) \centernot\implies \textsc{q}_{\mathbf{b0}}(\mathbf{b_0}) + \textsc{q}_{\mathbf{b1}}(\mathbf{b_1}) + \dots = \textsc{q}_{\mathbf{s}}(\mathbf{s})
$$
The solution passes through a requantization step similar to what is shown in the Quantization/Activation and BatchNormalization operators: one of the input branches (e.g., $\mathbf{b0}$) is chosen as reference ($\mathbb{Z}_\mathbf{s} \doteq \mathbb{Z}_\mathbf{b0}$) and as a consequence,
\begin{equation}
	\textsc{q}_{\mathbf{s}}(\mathbf{s}) = \textsc{q}_{\mathbf{s}}(\mathbf{b_0}) + \textsc{rq}_{(\mathbb{Z}_\mathbf{b1}\rightarrow\mathbb{Z}_\mathbf{s},D_1)}\big(\textsc{q}_\mathbf{b1}(\mathbf{b_1}) \big) + \textsc{rq}_{(\mathbb{Z}_\mathbf{b2}\rightarrow\mathbb{Z}_\mathbf{s},D_2)}\big(\textsc{q}_\mathbf{b2}(\mathbf{b_2}) \big) + \dots
	\label{eq:requantized_add}
\end{equation}

\begin{tcolorbox}[colback=green!5!white,colframe=green!75!black,title=In \nemo{}...]
To correctly represent Adds in the \textit{IntegerDeployable} representation, the network must be instantiating the \texttt{nemo.quant.pact.PACT\_IntegerAdd} modules.
Currently, instantiating this module is one of the few manual modifications required to a network's definition.
This is because the normal way of doing this in PyTorch (just using a \texttt{+}) does not instantiate a \texttt{torch.nn.Module} that can be augmented by \nemo{}.

In all modes except for \textit{IntegerDeployable}, \texttt{nemo.quant.pact.PACT\_IntegerAdd} behaves like a regular addition.
In  \textit{IntegerDeployable}, it performs requantization as shown in Eq.~\ref{eq:requantized_add}.
The $D$ is set through a \texttt{requantization\_factor} that defaults to 256, working in the same way as the one described in Section~\ref{sec:requantization} (i.e., it defaults to a relative requantization error $<1/256$).
\end{tcolorbox}

Note that if the paths diverge from an operator that is \textit{not} the final Quantization/Activation of a canonical layer, some of the operations explained in this document (e.g. BN folding) might be more complex and require additional work.
See for example Palossi~et~al.~\cite{PalossiOpenSourceOpen2019} for further details on this issue.

\begin{tcolorbox}[colback=green!5!white,colframe=green!75!black,title=In \nemo{}...]
The \texttt{nemo.transform.fold\_bn} has experimental support for inverse folding as explained in Palossi~et~al.~\cite{PalossiOpenSourceOpen2019}.
However, the strategy of branching from a non-Quantization/Activation operator is suboptimal and not recommended for networks that are meant to be quantized.
\end{tcolorbox}

\subsection{Pooling operators}
Max-Pooling is not touched by quantization, because all quantization mechanisms preserve relative ordering.
Therefore,
$$
\mathbf{t}[i] \geq \mathbf{t}[j] \;\iff\; \textsc{q}_\mathbf{t}(\mathbf{t})[i] \geq \textsc{q}_\mathbf{t}(\mathbf{t})[j]
$$

Average-Pooling, on the other hand, involves an implicit division by a factor $K_1\cdot K_2$ (the product of the pooling filter sizes), which could break the assumptions on integer images.
For this reason, a requantization-like operation is necessary. 
To do that, we transform the division in a product by $1/(K_1K_2)$, then we approximate it:
\begin{equation*}
	\frac{1}{K_1\cdot K_2} \approx \left\lfloor\frac{2^d}{K_1\cdot K_2}\right\rfloor \gg d
\end{equation*}
Therefore,
\begin{equation}
	\textsc{q}_\mathbf{p}(\mathbf{p}) = \left(\left\lfloor \frac{2^d}{K_1\cdot K_2} \right\rfloor \cdot \sum_{K_1,K_2} \textsc{q}_\mathbf{t}(\mathbf{t})\right) \gg d
	\label{eq:requantized_pooling}
\end{equation}

\begin{tcolorbox}[colback=green!5!white,colframe=green!75!black,title=In \nemo{}...]
In \textit{IntegerDeployable} representation, the \texttt{torch.nn.AvgPool2d} operators are transformed into \texttt{nemo.quant.pact.PACT\_IntegerAvgPool2d}.
These operators perform pooling as defined in Eq.~\ref{eq:requantized_pooling}.
\end{tcolorbox}

\subsection{Input representation}
The rules defined in our model enable propagating quanta in the network graph from each node representing an operation to its successors.
However, they leave out one question: what is the representation of the input of the network?
Often, input is naturally quantized (e.g., coming from an image with 8-bit channels, from analog-to-digital conversion, etc.) -- when the input has no obvious quantized representation, it has to be converted in an appropriate quantized version.

If the input has a representation similar to that of other activations in the network, i.e., with $\alpha=0$, then the model as described before directly applies to it, too.
However, it is possible that the ``natural'' representation of input has $\alpha\neq 0$.
In these cases, one possible approach is to add a bias to the first Linear node so that the input representation can be translated to the canonical $[0,\beta)$ one.

\begin{tcolorbox}[colback=green!5!white,colframe=green!75!black,title=In \nemo{}...]
It is possible to perform this operation to a network \texttt{net} using the \texttt{net.add\_input\_bias()} method.
\end{tcolorbox}

\subsection{Other operators}
There are many ``exotic'' operators that are not considered in this text (and not supported in \nemo{}).
For most of them, what is described here can be directly applied with minimal changes.
However, a particular mention is necessary for point-wise nonlinearities: most of these are used as alternative activation functions instead of ReLU.
Some of them can be integrated in the quantization/activation functions, often as thresholds.
Others, especially ones very sensitive in terms of dynamic range (e.g. exponentials) require switching back to real-valued (float) tensors to be applied.

\section*{Acknowledgement}
\nemo{} is an outcome of the European Commission Horizon 2020 ALOHA Project, funded under the EU's Horizon 2020 Research and Innovation Programme, grant agreement no. 780788.
The author also wants to thank Manuele Rusci, Alessandro Capotondi and Matteo Spallanzani for the many discussions that resulted in this technical report.
Thanks also to Marcello Zanghieri for proof-reading the first draft of this text.

\bibliographystyle{unsrt}
%\bibliography{template}

\end{document}